%% file: main.tex
\documentclass[letterpaper, 10 pt, conference]{ieeeconf}  
\IEEEoverridecommandlockouts                              
\usepackage[usenames, dvipsnames]{color}
\usepackage[noadjust]{cite}
\usepackage{booktabs}
\usepackage{graphicx}
\usepackage{subcaption}
\usepackage{float}
\usepackage{amsmath}
\usepackage{algorithmicx}
\usepackage{multirow}
\usepackage{mathrsfs}
\graphicspath{{images}{../images/}}
\usepackage{graphicx}
\usepackage[utf8]{inputenc}
\usepackage{amsmath}
\usepackage[hidelinks]{hyperref}

\usepackage[usenames,dvipsnames,table]{xcolor}
\usepackage{soul,xcolor}
\usepackage{siunitx}
\usepackage[export]{adjustbox}
\usepackage{multicol}
\usepackage{tikz}

\title{\LARGE \bf
Vehicle Motion Forecasting using Prior Information and Semantic-assisted Occupancy Grid Maps\\
}

\author{Rabbia Asghar$^{1}$, Manuel Diaz-Zapata$^{1,2}$, Lukas Rummelhard$^{1}$, Anne Spalanzani$^{1}$, Christian Laugier$^{1}$\\
\thanks{$^{1}$~Univ. Grenoble Alpes, Inria, 38000 Grenoble, France, email:  FirstName.LastName@inria.fr $^{2}$~Univ Lyon, Inria, INSA Lyon, CITI Lab}%
}

\begin{document}
\setstcolor{red}

\maketitle
\thispagestyle{empty}
\pagestyle{empty}
\setlength{\belowdisplayskip}{2pt}
\widowpenalty10000
\clubpenalty10000
\addtolength{\abovedisplayskip}{-5pt}

\begin{abstract}
Motion prediction is a challenging task for autonomous vehicles due to uncertainty in the sensor data, the non-deterministic nature of future, and complex behavior of agents.
In this paper, we tackle this problem by representing the scene as dynamic occupancy grid maps (DOGMs), associating semantic labels to the occupied cells and incorporating map information.
We propose a novel framework that combines deep-learning-based spatio-temporal and probabilistic approaches to predict vehicle behaviors.
Contrary to the conventional OGM prediction methods, evaluation of our work is conducted against the ground truth annotations. We experiment and validate our results on real-world NuScenes dataset and show that our model shows superior ability to predict both static and dynamic vehicles compared to OGM predictions. Furthermore, we perform an ablation study and assess the role of semantic labels and map in the architecture.
\end{abstract}

\input{sections/introduction}
\input{sections/related_work}

\input{sections/approach}
\input{sections/experiments}
\input{sections/results.tex}
\input{sections/conclusion}




%
%

\bibliographystyle{IEEEtran}
\bibliography{main}%

\end{document}

%% file: sections/introduction.tex
\section{INTRODUCTION} \label{sec:introduction}
If autonomous vehicles are to effectively increase road safety \cite{nhtsa}, their situational awareness should reach and overcome human driver capabilities.
With the advancements in deep learning and computer vision, the research has significantly progressed in this domain to improve perception of the scene and predict complex interaction of agents.
To advance on their actual deployment, and to meet the requirements of such critical embedded devices, 
a balance between explainable artificial intelligence methods and interpretable deep-learning approaches is to be found.

In line with this, dynamic occupancy grid maps (DOGMs) provide a simplified discrete representation of the scene in bird's-eye-view which can be effectively integrated with deep learning methods to learn complex behaviours. 
The OGMs offer efficient sensor fusion, real-time computation and dense probabilistic occupancy representation, allowing for modeling sensor and prediction uncertainties. 
In our work, we consider Bayesian-filter-based DOGMs \cite{lukas22} to represent the scene with respect to a fixed reference frame, enabling long-term occupancy predictions.

\begin{figure}[ht]
	\centering
	    \includegraphics[trim={0 0 0 0},clip,width=0.9\columnwidth]{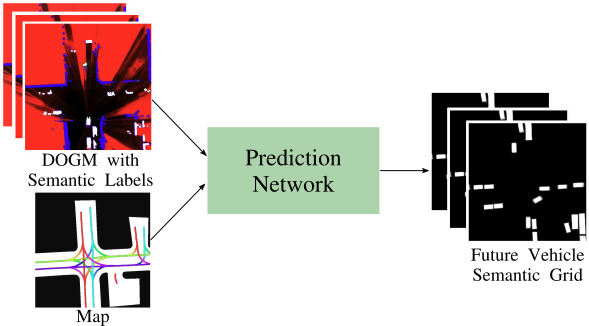}
\caption{\small Our proposed architecture. Vehicle semantic labels are associated to DOGM occupied cells. Sequence of these DOGMs and map information are fed into the network to predict a sequence of future vehicle semantic grids. }
\label{fig:overview-front}
\vspace{-0.5cm}
\end{figure}

These grid predictions, however, present challenges of blurry dynamic vehicles and their evaluation.
With few common benchmark and direct comparison methods being available, the future prediction of OGMs are mostly evaluated against the actual generated grid  (\cite{Toyungyernsub2022}, \cite{asghar22}) rather than the ground truth.
This carries the inherent risk of overlooking potential errors in OGMs generation during training and validation.
To address these concerns, we train our network against ground truth annotations to predict vehicle semantic grids. Consequently, this also allows the network to focus on agents and address the problem of blurry vehicles prediction.

To further improve the prediction of vehicle behaviors in the scene, we incorporate contextual information in input that relies on vehicle semantics and prior map information.
We associate semantic labels to the occupancies in DOGM input and also include corresponding rasterized map, see Fig. \ref{fig:overview-front}. 

The main contributions of this paper is a novel framework that incorporates sequences of DOGMs along with semantic labels and maps to predict vehicle motion as semantic grids. A spatio-temporal module in the network captures the evolution of scene while a probabilistic model aids in learning diverse future predictions. Instead of comparing predictions against the actual generated grids, this framework permits us to evaluate our results against ground truth annotations. We compare our results against conventional OGM methods on the real-world NuScenes dataset \cite{nuscenes2019} and show that our model has superior ability to predict vehicle motion in the scene. Furthermore, we perform an ablation study and report the significance of various input components in our network.
\begin{figure*}[ht]
	\centering
	    \includegraphics[trim={0 0 0 0},clip,width=1.9\columnwidth]{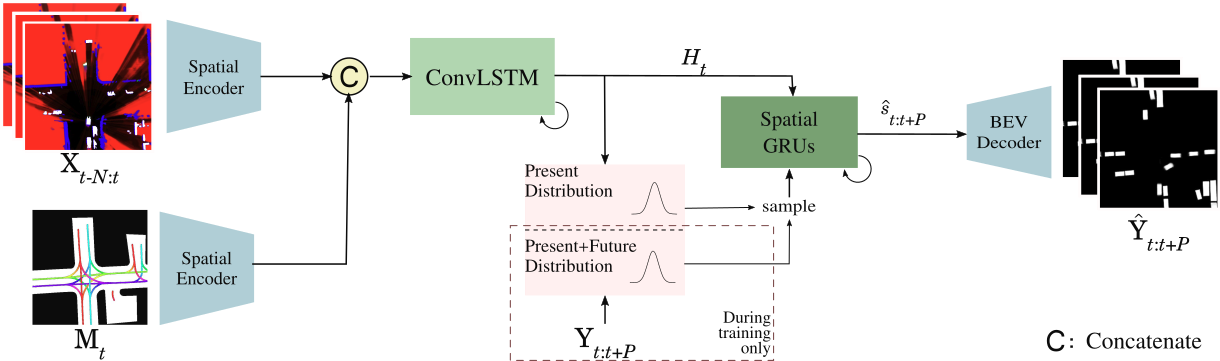}
\caption{\small An overview of our proposed network. DOGM with semantic labels and raster map features are fed to spatio-temporal and the probabilistic blocks to capture the evolution of the scene and predict a sequence of future vehicle semantic grids. }
\label{fig:overview}
\vspace{-0.5cm}
\end{figure*}

%% file: sections/related_work.tex
\section{Related Work} \label{sec:related_work}
\subsection{Occupancy Grid Prediction}
Recent works have considered the problem of future ego-centric OGM predictions (\cite{8500567},  \cite{Dequaire2017}, \cite{mohajerin2019multi}, \cite{schreiber2019long}, \cite{lange2020attention}, \cite{Toyungyernsub2022})
by incorporating spatio-temporal deep-learning methods, involving combinations of convolutional neural networks (CNNs) and recurrent neural networks (RNNs).
These works predict the complete scene as OGMs and encounter similar challenges for  forecasting such as blurriness, loss of scene structure and disappearance of dynamic agents. Asghar \textit{et al.} \cite{asghar22} proposed making future DOGM prediction in a fixed reference frame by representing the scene as an allo-centric grid. This framework helps in preserving scene integrity especially when the ego-vehicle turns.

Most future OGM prediction methods are evaluated against their own OGMs, that are considered ground truth. 
Schreiber \textit{et al.} \cite{schreiber2019long} further process their DOGMs to generate the static and dynamic labels for evaluation. Mann \textit{et al.} \cite{mann2022predicting} overlayed semantic vehicle labels to their DOGM input but only evaluated against the modified DOGM grids.

In this paper, contrary to the conventional OGM prediction methods, we predict the vehicle occupancy grid and evaluate it against the annotated ground truth. We carry forward the work of Asghar \textit{et al.} \cite{asghar22} and incorporate additional information such as map and vehicle semantics and show how it can improve the prediction capabilities of vehicles in the scene. Moreover, we incorporate probability distributions in our network to consider diverse future predictions.
 
\subsection{Motion forecasting}
In autonomous driving literature, most state-of-the-art works rely on the availability of annotated and heavily processed data
(\cite{mahjourian2022occupancy}, \cite{gao2020vectornet})
on motion forecasting. 
Other traditional methods address this problem with detection, tracking and prediction modules
and run the risk of missing out agents due to confidence thresholding. 
Instead, many end-to-end methods address jointly fusion of these modules 
(\cite{liang2020pnpnet}, \cite{wu2020motionnet}).
 Among these approaches, Fiery \cite{fiery2021}, MP3 \cite{casas2021mp3} and FishingNet \cite{hendy2020fishing} predict future occupancy grids with semantic labels without tracking.
Different from these methods, we use DOGMs that offer real-time occupancy information of the scene in bird’s-eye-view (BEV), provide probabilistic information where only partial information is available and allow fusion of different sensors, as well as different configurations. To model probabilistic future predictions, we incorporate the conditional variational approach proposed in Fiery \cite{fiery2021}.

%% file: sections/approach.tex
\section{System Overview} \label{sec:approach}
We discuss in detail here the proposed approach for prediction, the pipeline is summarized in Fig. \ref{fig:overview}.

\subsection{Dynamic Occupancy Grid Maps}\label{sec:approach-a}
Dynamic occupancy grid maps provide a BEV grid representation of the environment, where each cell carries information about the associated occupancy, as well as its dynamics. A key characteristic of OGMs is that each cell state is always estimated, independent of object detection or tracking system, limiting thresholding-induced risks.

In our work, we incorporate Bayesian filter based Conditional Monte Carlo Dense Occupancy Tracker (CMCDOT) \cite{lukas22} to generate DOGMs from lidar. This approach associates 4 occupancy states to each cell, carrying the probabilities of being i) free, ii) occupied and static, iii) occupied and dynamic and iv) if the occupancy is unknown. This framework allows real-time generation of the grids on real vehicles.

Following the work of Asghar \textit{et al.} \cite{asghar22}, we process and represent the grid in a fixed reference frame, such that in a sequence of grids, it is the ego-vehicle that displaces and the static scene remains fixed. This approach leverages the odometry and ego-centric observations to generate an allo-centric representation. 
We incorporate three of the DOGM states and represent them as RGB images, represented in Fig. \ref{subfig:sem2}. The Red channel represents the unknown state, dynamic and static occupied states are assigned Green and Blue channel respectively. Absence of all three colours implies free space.

\subsection{Inclusion of semantic labels}\label{subsec:semantics}
The conventional DOGM methods observe space around the ego-vehicle and provide occupancy information without any semantics, such as whether the occupancy belongs to a vehicle or to some road barrier. In this work, we assume the availability of vehicle semantic information for all cells that are occupied, whether static or dynamic. 
Possible sources to obtain this information may involve exploiting vehicle semantics from object detection methods such as PointPillars \cite{lang2019pointpillars}, or by fusing camera-image-based semantic segmentation information to the DOGM \cite{erkent2018semantic}. 

\begin{figure} [h]
    \captionsetup[subfigure]{justification=centering}
    \centering
    \begin{subfigure}[c]{0.3\columnwidth}
        \centering
        \includegraphics[width=\columnwidth]{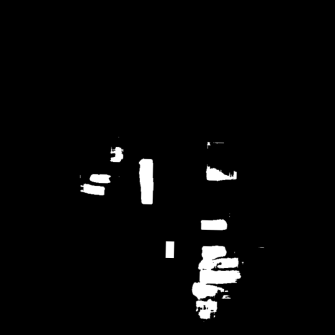}
        \caption[]{{\small Semantic information source }}    
        \label{subfig:sem1}
    \end{subfigure}
    \hfill
    \begin{subfigure}[c]{0.3\columnwidth}  
        \centering 
        \includegraphics[width=\columnwidth]{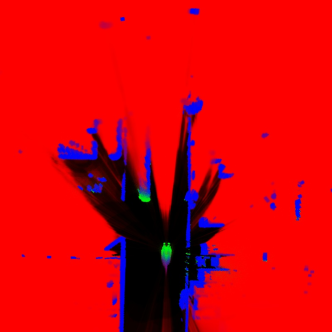}
        \caption[]{\small DOGM state grid}
        \label{subfig:sem2}
    \end{subfigure}
    \hfill
    \begin{subfigure}[c]{0.3\columnwidth} 
        \centering 
        \includegraphics[width=\columnwidth]{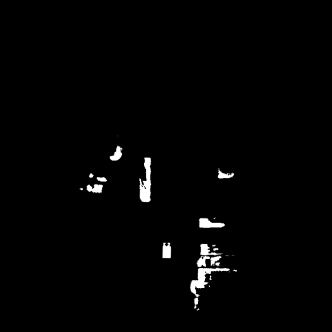}
        \caption[]{{\small Associated semantic labels}}
        \label{subfig:sem3}
    \end{subfigure}
    \caption[]
    {\small Inclusion of vehicle semantic labels. (a) vehicle semantics predicted by PointPillars \cite{lang2019pointpillars}, (b) DOGM grid-image, (c) binary vehicle semantic labels assigned to the occupied cells in DOGM.}
    \label{fig:semantics}
\vspace{-0.6cm}
\end{figure}

Figure \ref{fig:semantics} illustrates the generation of vehicle labels for respective DOGM grid. If a cell is occupied (represented in blue or green in \ref{subfig:sem2}), and if it represents a vehicle in the semantic information (\ref{subfig:sem1}), it is assigned as a vehicle semantic label to the occupancy in the DOGM. A binary semantic labels grid (\ref{subfig:sem3}) is generated for every DOGM, where the cells are set to 1 if they are occupied and belong to the vehicles.
This  grid is concatenated to the DOGM input as another channel.

\subsection{Problem Formulation}\label{sec:approach-b}
We formally define the task of predicting the vehicle motion, see Fig. \ref{fig:overview}.
Let $X_t \in \mathrm{R}^{4 \textrm{x} w \textrm{x} h}$ be the $t$-th frame of the grid-image where $w$ and $h$ denote the  width and height respectively. $X_t$ comprises of three occupancy state channels (Fig. \ref{subfig:sem2}) and  one semantic grid channel (Fig. \ref{subfig:sem3}).
Let $M \in \mathrm{R}^{3 \textrm{x} w \textrm{x} h}$ be the BEV rasterized map of the scene, encoding the driveable space, lane structures and respective directions of traffic, and $Y_t \in \mathrm{R}^{3 \textrm{x} w \textrm{x} h}$ be the $t$-th frame of BEV vehicle semantic grid.

Given a set of input sequence $X_{t-N:t}$ and the corresponding map $M_t$, the task of the network is to predict a sequence of future vehicle grids $\hat{Y}_{t:t+{P}}$. 
The target and predicted output sequences are denoted by ${Y}_{t:t+{P}}$ and $\hat{Y}_{t:t+{P}}$ respectively, where $P$ is the prediction horizon.

\subsection{Prediction Architectures}\label{subsec:NN-architectures}
Our proposed architecture, Fig. \ref{fig:overview}, takes inspiration from both spatio-temporal and conditional variational approach. 
To extract spatial features from input sequence, the input grid $X_{t-N:t}$ is successively fed to the spatial encoder with 2D-convolutions. An identical spatial encoder extracts map features from an RGB rasterized map image $M_t$. These two features are concatenated and sequentially fed to ConvLSTM (Convolutional Long Short Term Memory) \cite{shi2015convolutional} block in time order. We employ 4 units of ConvLSTM to capture spatio-temporal features of the complete input sequence.
The output state, $H_t$, from this block is then fed into the probabilistic distributions and spatial GRUs (Gated Recurrent Units).

Given a set of input states, the scene can evolve in multiple possible ways. To take this into account, we incorporate the present and future distribution as proposed by Hu \textit{et al.} (\cite{fiery2021}, \cite{hu2020probabilistic}). 
During training, the \textit{present}$+$\textit{future distribution} learns to model the distribution conditioned on $H_t$ and ground truth labels ${Y}_{t:t+{P}}$, while during inference, the \textit{present distribution} models the future states solely from $H_t$. The two distributions are parameterized to be diagonal Gaussian distributions with 32 latent space dimensions.
Kullback-Leibler divergence loss, $\mathcal{L}$$_{KL}$, is introduced to encourage \textit{present distribution} to match the \textit{present}$+$\textit{future distribution}.

The spatial GRUs take the ConvLSTM output, $H_t$, and recursively predict future states, $\hat{s}_{t:t+P}$,  depending on the sample from the probabilistic distributions. 
We follow the implementation by \cite{fiery2021}, that comprises of 3 units of convolutional GRUs \cite{ballas2015delving} for every future timestep prediction.

The future state predictions $\hat{s}_{t:t+P}$ are then fed to a frame decoder with 2D-DeConvolutions to generate BEV frames, $\hat{Y}_{t:t+{P}}$. The vehicle semantic grids are trained with binary cross-entropy loss $\mathcal{L}$$_{BCE}$ with a positive sample weight of 5. The loss function for our model is defined as

\begin{align*}
\mathcal{L} =  \lambda_{b} \mathcal{L}_{BCE} +  \lambda_{k} \mathcal{L}_{KL}    
\end{align*} \label{eq:loss}

where $\lambda_{b}$ and $\lambda_{k}$ are the respective loss weights. 

%% file: sections/experiments.tex
\section{Experiments} \label{sec:implementation}

\subsection{Dataset}\label{subsec:experiments}
We consider the real-world NuScenes dataset \cite{nuscenes2019} to study the prediction performance of our proposed model.
The original dataset consists of 700 scenes for training and 150 scenes for validation; 
each scene duration is approximately 20s.
We generate the Lidar based DOGMs with available odometry. Each grid sequence is defined with respect to a fixed reference frame and starts with the ego-vehicle heading facing up, capturing the scene 10m behind and 50m ahead of it. This allo-centric grid dimension is fixed to 60x60m, with a resolution of 0.1m per cell. 
For the same grid configuration, raster maps and the binary vehicle semantic grids are generated. 
The raster map contains the driveable lanes information available in the dataset while the vehicle grid contain the projection of ground truth annotation boxes in BEV. 
Vehicle semantic grids only contain the vehicles that were \textit{perceived} by the DOGM in the input sequence and do not consider new vehicles entering the scene, or the vehicles that are entirely unobserved. 
A vehicle is considered \textit{perceived} if any DOGM cell in that space has the probability higher than 0.3 that it is occupied. For predictions, we also include the ego-vehicle box annotation as its motion may influence the behaviour of nearby vehicles.

The semantic labels for DOGMs are generated by comparing the DOGM grid with available semantic information and only for the cells where the occupancy probability of either static or dynamic state of DOGM is higher than 0.3. Two sources of semantic information are considered in experiments:  i) vehicle annotations in the dataset ii) vehicle semantics predicted by the Lidar based PointPillars network \cite{lang2019pointpillars}.

Each sequence in our dataset is unique and has a time duration of 3.5s.
In total, we have 3,499 training and 750 validation sequences, respectively.

\begin{figure*}[ht]
	\centering
	    \includegraphics[trim={0 0 0 0},clip,width=1.5\columnwidth]{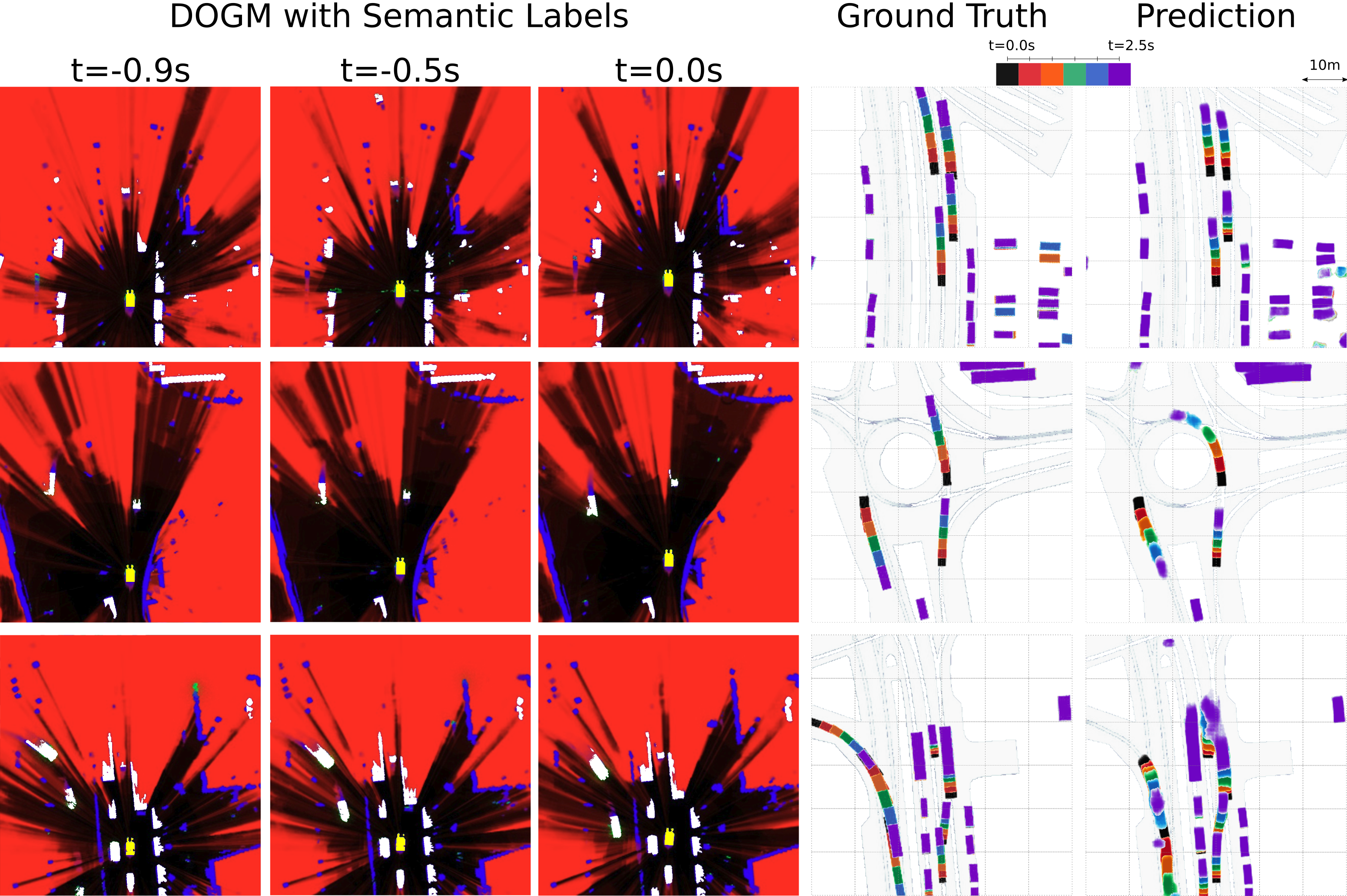}
\caption{\small Vehicle occupancy prediction examples on three scenes from the Nuscenes Datatset \cite{nuscenes2019}, covering 60x60m. Left three cloumns display the DOGM input with semantic labels for 1s duration. Semantic labels are indicated in white while ego-vehicle is shown in yellow. Ground truth and occupancy predictions are overlayed on map and are colour coded to visualize range of 0.0s to 2.5s from black to purple. Intensity of colours in prediction represents respective probability.}
\label{fig:quality-results}
\vspace{-0.5cm}
\end{figure*}

\subsection{Training}\label{subsec:training}
The input sequence $X_{t-{9}:t}$ consists of 10 frames (1.0s) and the corresponding raster map. The network is trained to make predictions $\hat{Y}_{t:t+{5}}$ for present and 5 future frames, for instances 0.0 to 2.5s, with steps of 0.5s.
For training, the grid images are resized to 256x256 pixels, thus each pixel has a resolution of approximately 0.23m. The model is trained using Adam Optimizer at the learning rate of $2$x${10}^{-4}$ for 85 epochs and the batch size of 6. The loss coefficients are set to  $\lambda_{b}=1$ and  $\lambda_{k} = 0.005$. 

\subsection{Training variations for evaluation}\label{subsec:training_b}

To generate semantic labels with the help of the PointPillars \cite{lang2019pointpillars}, the respective network was trained on the NuScenes dataset to predict the vehicle grid segmentation from Lidar data. Compared to our dataset, this was trained on the same grid dimension and resolution (60x60m at 0.1m/pixel) but in an egocentric reference frame with the vehicle fixed in the center. Note that the ego-centric grid space covered under this does not completely match the space covered by our allo-centric grid dataset.
The model's vehicle semantics intersection over union (IoU) on the validation set was 0.53 with the precision of 0.77.
The PointPillars semantics, therefore, offers evaluation of our approach on a relatively noisy source of semantic labels.
The results were processed to associate semantic labels to the occupied cells in the DOGM sequence, as described in sec. \ref{subsec:semantics}.

In conventional OGMs future prediction approaches, the models are trained to take OGM input sequence and predict a sequence of future OGMs. To compare the proposed approach against these methods, we follow the prediction setup in Asghar \textit{et al.} \cite{asghar22} and evaluate against two video prediction methods (LMC-memory \cite{lee2021video} and PredRNN \cite{wang2022predrnn}) and the TAAConvLSTM  network proposed by Lange \textit{et al.} \cite{lange2020attention}.
The networks were re-trained for the DOGM dataset described in \ref{subsec:experiments}.
The image size was resized to 192x192 pixels for our network, LMC-Memory, PredRNN and 160x160 pixels for TAAConvLSTM respectively. All networks, including ours, were trained for 30 epochs.
In an effort to keep it as similar as possible, our network was trained and evaluated for all vehicles annotated in the dataset, including any new agents that may enter the scene in the future.


For ablation studies, we train the network with different input configurations with the training set up described in section \ref{subsec:training}.


%% file: sections/results.tex
\section{Results} \label{sec:results}

\subsection{Qualitative Evaluation}\label{subsec:qltyevaluation}
The qualitative prediction results in Fig. \ref{fig:quality-results} depict various driving scenes with predictions based on the mean values of the \textit{present distribution}. Top row displays four dynamic vehicles with seemingly similar traffic direction and numerous static vehicles on both sides of the road. The availability of the semantic labels for very small or patchy cluster of occupied cells in the DOGM help the network learn the occupancy of static vehicles. 
The middle row shows three vehicles around the roundabout. While one of the vehicles originally exits the roundabout, prediction shows another possible outcome and continues on it. This prediction is particularly aided by providing the map information in the input as the roundabout layout is not perceived by the DOGM.
The bottom row captures diverse behaviours on a 3-lane highway. A vehicle just enters the scene (from the bottom of the grids), but decelerates to keep a safe distance from the static vehicle in front. The ego-vehicle in the middle lane can be seen changing lane, followed by another vehicle. In the same scene the vehicles ahead, however, struggle to retain sharpness of their predictions due to their close proximity towards the end. We propose that the instance segmentation of vehicles can help improve these results.
In all three scenes, there are many cells in the DOGMs that show occupancy and could be misinterpreted as vehicles. Thanks to the associated semantics labels, the network can reliably predict both static and dynamic vehicles.

\subsection{Quantitative Evaluation}\label{subsec:qntyevaluation}

\subsubsection{Comparison with the OGM prediction literature}\label{subsec:OGMevaluation}
Common challenges in the future prediction of OGMs include the blurriness of static components and the vanishing of dynamic agents. We evaluate here the forecasting of static and dynamic vehicles in our proposed framework against the state-of-the-art OGM approaches. Two different sources of semantic information are compared within our approach:  labeled annotations and Pointpillar-based semantics.

\begin{figure}
    \centering
    \hfill
    \begin{subfigure}[c]{0.35\columnwidth}
        \centering
        \includegraphics[width=\columnwidth]{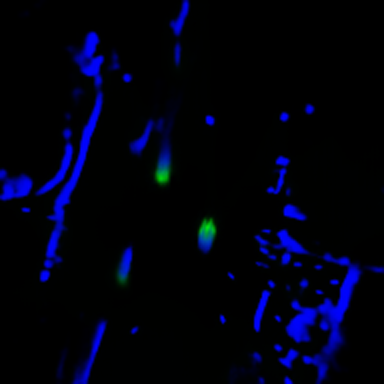}
        \caption[]{{\small Predicted grid}}    
        \label{subfig:pred_samp}
    \end{subfigure}
    \hfill
    \begin{subfigure}[c]{0.35\columnwidth}  
        \centering 
        \includegraphics[width=\columnwidth]{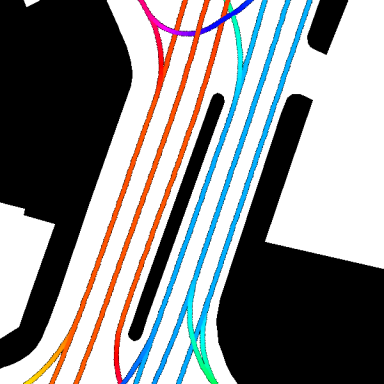}
        \caption[]{{\small  Driveable space}}    
        \label{subfig:map}
    \end{subfigure}
    \hfill
    \hfill
    \vskip\baselineskip
    \hfill
    \begin{subfigure}[c]{0.35\columnwidth}   
        \centering 
        \includegraphics[width=\columnwidth]{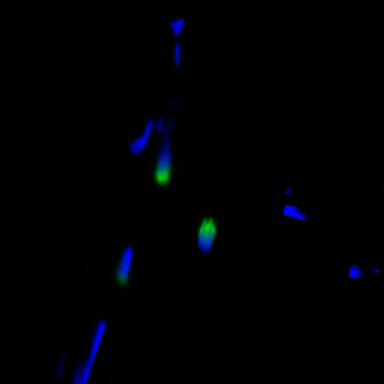}
        \caption[]{{\small Vehicle prediction}}
        \label{subfig:pred_veh}
    \end{subfigure}
    \hfill
    \begin{subfigure}[c]{0.35\columnwidth}   
        \centering 
        \includegraphics[width=\columnwidth]{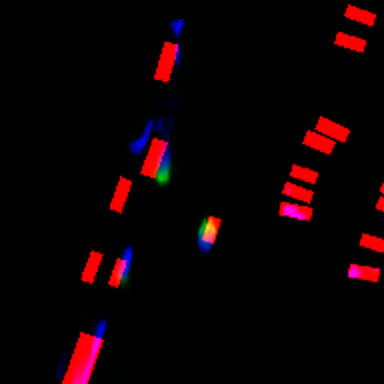}
        \caption[]{{\small Overlayed GT}}
        \label{subfig:gt}
    \end{subfigure}
    \hfill
    \hfill
    \caption[ ]
    {\small Preparation for vehicle soft IoU computation for an OGM prediction instant (a). With help of prior map (b), vehicle predictions are isolated (c). The ground truth (GT) vehicle boxes in red (d) are overlayed on the grid, with intersection illustrated in yellow or pink.} 
    \label{fig:OGM-eval}
\vspace{-0.6cm}
\end{figure}

Since OGMs capture occupancy states of the whole scene, their prediction results are processed to eliminate all possible occupancies that do not represent vehicles. 
To achieve this, the map and ground truth annotations available in the dataset are exploited. 
All occupied cells that are not on driveable lanes or that represent other categories such as pedestrian or moveable barriers, are removed.
Figure \ref{fig:OGM-eval} illustrates a prediction example, before and after the elimination of the scene; leaving behind only vehicle predictions. Overall, to give benefit to the OGM predictions, all occupied cells that are not within 2 meters of the annotated ground truth are also removed.


 Soft Intersection over Union (soft-IoU) metric is employed at 5 time steps in future ($t$= 0.5,...,2.5s). Contrary to the more popular IoU metric for semantic segmentation, soft-IoU does not evaluate binary classification but considers all prediction values in range 0 to 1. Soft-IoU at a time step is computed as follows 
\begin{align} \label{eq:soft-IoU}
soft - IoU =  {\frac{{\sum \nolimits _{i \in x,y} {{p_{i}} \cdot p_{i}^*} }}
{{\sum \nolimits _{i \in x,y} {{p_{i}+ p_{i}^* - {p_{i}} \cdot p_{i}^*}} }}}
\end{align}
where $p_{i}$ represents predicted occupancy of a pixel while $p_{i}^*$ denotes the ground truth occupancy, which is either 1 or 0.

We see in Table \ref{table:results-b} that the soft-IoUs of our proposed approach, even with the noisy source of Pointpillars-based semantics, are significantly higher than the ones in case of OGM predictions. 
Low soft-IoU scores of the OGM networks signify a key characteristic of these methods that tends to predict only part of the vehicle; the remaining part is either unknown (incomplete information) or free (incorrect information).

\begin{table}[h]
\centering
\begin{tabular}{c|c|c|c|c|c|}
\cline{1-5} 
\hline
\multicolumn{1}{c}{\textbf{Network}} & \multicolumn{2}{c}{\textbf{Retention of Vehicles(\%)}}  & \multicolumn{2}{c}{\textbf{soft-IoU}$(\uparrow)$}\\\hline
\cline{1-5} 
\multicolumn{1}{c}{\textbf{}} & \multicolumn{1}{c}{\textbf{Dynamic}}   & \multicolumn{1}{c}{\textbf{Static}}  & \multicolumn{2}{c}{}\\
\cline{2-3} 
\multicolumn{1}{c}{LMC-Memory \cite{lee2021video}} & \multicolumn{1}{c}{72.21}  & \multicolumn{1}{c}{86.75}  & \multicolumn{2}{c}{0.183} \\
\multicolumn{1}{c}{PredRNN \cite{wang2022predrnn}} & \multicolumn{1}{c}{{70.63}}  & \multicolumn{1}{c}{88.18}  & \multicolumn{2}{c}{{0.159}} \\
\multicolumn{1}{c}{TAAConvLSTM \cite{lange2020attention}} & \multicolumn{1}{c}{{60.84}}   & \multicolumn{1}{c}{82.18} & \multicolumn{2}{c}{{0.174}} \\ \cline{1-5} 
\multicolumn{1}{c}{Ours+PointPillars} \cite{lang2019pointpillars}  & \multicolumn{1}{c}{{82.98}} & \multicolumn{1}{c}{88.50}  & \multicolumn{2}{c}{0.360} \\ 
\multicolumn{1}{c}{Ours}  & \multicolumn{1}{c}{\textbf{85.94}} & \multicolumn{1}{c}{\textbf{98.62}}  & \multicolumn{2}{c}{\textbf{0.450}} \\ \hline
\end{tabular}
\caption{\small Comparison of vehicle predictions against conventional OGM prediction approaches based on retention of vehicles in the scene and soft-IoU during the 2.5s prediction horizon. Ours+PointPillars model incorporates PointPillars-based semantic information. Bold indicates best.} 
\label{table:results-b}
\vspace{-0.5cm}
\end{table}

\begin{figure}[h]
\vspace{-0.5cm}
	\centering
	    \includegraphics[trim={0 0 0 0},clip,width=0.9\columnwidth]{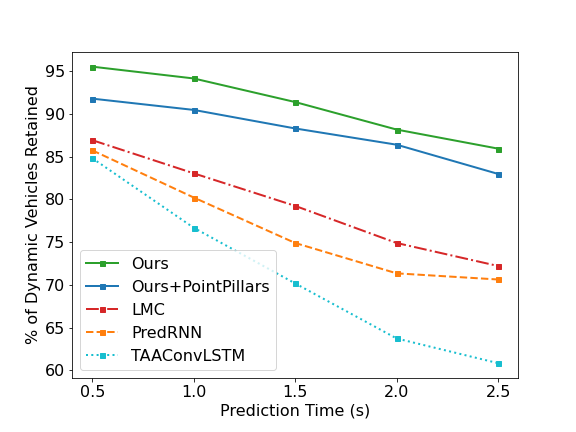}
\caption{\small Retention of dynamic vehicles in the scene.}
\label{fig:iou-plot}
\vspace{-0.4cm}
\end{figure}


We further evaluate against the OGM predictions on their ability to retain vehicles up to a 2.5s prediction horizon.  
Since original OGMs tend to partially predict agents, instead of summing the overlap in soft-IoU, we consider overlap of ground truth with {any} part of a vehicle prediction to be retention of the given vehicle.
 For this, we note the vehicles \textit{perceived} in the input sequence and
 track each vehicle in the scene to compare if {any} cell in the prediction grid with probability greater than 0.1 overlaps the ground truth.
In our validation set, excluding ego-vehicle, a total of 5,288 vehicles were completely or partially \textit{perceived} in the input sequence, and among those 1,522 were dynamic. 

Table \ref{table:results-b} compares the percentage of dynamic and static vehicles that the network was able to retain by 2.5s prediction horizon. While the OGM predictions show a good tendency to retain static vehicles, the dynamic vehicle retention remain below 73\% for the three studied networks. Our proposed method in comparison is able to retain most of the static vehicles and up to 83\% of the dynamic vehicles with the noisy Pointpillars semantics. Fig. \ref{fig:iou-plot} plots the percentage of dynamic vehicles retained at different timesteps. Note that the OGM methods lose around 15\% of the dynamic vehicles in the first 0.5s. We conjecture that these are the dynamic vehicle that are observed only for a brief time and on sparse cells during the input sequence. Our network has the advantage of semantic labels associated to those intermittent occupancies and is able to hold on to those predictions.

\subsubsection{Ablations study}\label{subsec:ablationevaluation}
We perform an ablation study on the various input configurations of our proposed network. Similar to binary classification evaluation metrics, the predicted semantic grids are evaluated with intersection over union (\textbf{IoU}) and the area under precision-recall curve (\textbf{AUC}). To compute IoU, the threshold is set to 0.5 while AUC is computed for 100 linearly spaced thresholds. A high AUC represents both high precision and recall, i.e., the proportion of vehicle predictions that were correct and proportion of actual occupancy that were correctly predicted respectively. 

\begin{table}[h]
\centering
\begin{tabular}{ccc}
\cline{1-3}
\hline
\multicolumn{1}{c}{\textbf{Input}} & {\textbf{IoU}$(\uparrow)$}  & {\textbf{AUC}$(\uparrow)$} \\
\hline
\multicolumn{1}{c}{DOGM} & {0.336}  & {0.496} \\
\multicolumn{1}{c}{DOGM + Map} & {0.430} & {0.604}  \\
\multicolumn{1}{c}{DOGM + Semantics} & {0.590} & {0.771}  \\
\multicolumn{1}{c}{{DOGM + Map + Semantics}} & \textbf{{0.605}} & \textbf{{0.781}}  \\ \hline
\end{tabular}
\caption{\small Ablation study on adding information to DOGM. We report mean values of IoU and AUC over 2.5s prediction horizon.}
\label{table:ablation-study}
\vspace{-0.4cm}
\end{table}


We compare the results with different combinations of network input:
i) the DOGM alone in the input, 
ii) DOGM along with the map,
iii) DOGM with vehicle semantic labels, and 
iv) finally all three together, the complete architecture as proposed.
For ablation study, vehicle annotations available in the dataset are considered for semantic information source.

Table \ref{table:ablation-study} shows the average values of IoU and AUC over 2.5s prediction horizon for the four cases.
 While the best performance is seen with our proposed network inputs, the results with \textit{DOGM+Semantics} are not too far behind. 
 The addition of semantic labels to the occupied cells plays a key role in improving the vehicle motion prediction.
Moreover, with the occupancy information available in DOGMs, the network can learn to estimate the road structure in the scene and, thus, diminishing the benefits of map inclusion in the input. 
Nevertheless, in absence of semantic labels, the map information does notably assist in improving prediction results. As it was also seen in the qualitative results, the availability of driveable space aids future predictions in the scene space that may not be perceived during the input.


%% file: sections/conclusion.tex
\section{Conclusion}\label{sec:conclusion}
In this work, we have proposed a novel framework that combines dynamic occupancy grid maps with semantic labels and offline map to predict future vehicle motion. 
Our network integrates traditional probabilistic DOGM approach with spatio-temporal and conditional variational deep learning methods, to learn and predict probabilistic vehicle behaviours. 
Evaluating our work against conventional OGM prediction methods, we report that our model shows superior ability to predict both static and dynamic vehicles. We also perform an ablation study on the proposed input framework
 and note that, more than the map information, the addition of semantic labels plays a significant role in improving the results. 
For future work, we envision
fusion of semantic labels in the DOGM grid generation
and prediction of other agents in the scene as well, such as pedestrians or cyclists.